\pdfoutput=1

\documentclass[11pt]{article}

\usepackage{acl}

\usepackage{times}
\usepackage{latexsym}

\usepackage[T1]{fontenc}

\usepackage[utf8]{inputenc}

\usepackage{microtype}

\usepackage{inconsolata}

\newcommand{\citeit}[1]{\textcolor{red}{\textit{[CITE]}}}

\usepackage{lipsum}
\usepackage{hhline}
\usepackage{array}
\usepackage{tabularx}
\usepackage{enumerate}
\usepackage{multirow}

\usepackage[none]{hyphenat}

\title{NuclearQA: A Human-Made Benchmark for Language Models for the Nuclear Domain}

\author{Anurag Acharya,~~ Sai Munikoti,~~ Aaron Hellinger,~~ Sara Smith, \\ \textbf{Sridevi Wagle,}~~ \textbf{Sameera Horawalavithana} \\
        Pacific Northwest National Laboratory \\ Richland, WA, USA \\
\texttt{\{anurag.acharya, sai.munikoti, aaron.hellinger, sara.smith,} \\ \texttt{sridevi.wagle, yasanka.horawalavithana\}@pnnl.gov}}

\begin{document}

\maketitle

\begin{abstract}

As LLMs have become increasingly popular, they have been used in almost every field. But as the application for LLMs expands from generic fields to narrow, focused science domains, there exists an ever-increasing gap in ways to evaluate their efficacy in those fields. For the benchmarks that do exist, a lot of them focus on questions that don't require proper understanding of the subject in question. In this paper, we present NuclearQA, a human-made benchmark of 100 questions to evaluate language models in the nuclear domain, consisting of a varying collection of questions that have been specifically designed by experts to test the abilities of language models. We detail our approach and show how the mix of several types of questions makes our benchmark uniquely capable of evaluating models in the nuclear domain. We also present our own evaluation metric for assessing LLM's performances due to the limitations of existing ones. Our experiments on state-of-the-art models suggest
that even the best LLMs perform less than satisfactorily on our benchmark, 
demonstrating the scientific knowledge gap of existing LLMs.

\end{abstract}

\section{Introduction}

With the current rapid advancement in the field of Large Language Models (LLMs), they have been increasingly used for a wide variety of tasks across several domains. Among them, one of the more popular domains in recent times has been the scientific domain~\cite{taylor2022galactica, cohan2020specter, beltagy2019scibert}. There have been several models that have aimed to tackle the difficult task of scientific reasoning and understanding, and the results have been mixed, with these models performing well in some cases but not in others. Unfortunately, our ability to evaluate these models have been less than ideal due to lack of proper benchmarks.

While there exists numerous benchmarks for the fields of general question answering, commonsense reasoning, and so on, most of these usually draw from existing resources that already exist, like popular trivia show questions, high school and college notebooks and text, online news, and so on. But even then, the focus is mostly on generic and broad topics that can be used by all types of models, creating a dearth of such benchmarks for narrow, specific yet highly important sub-fields. Additionally, even when such benchmarks are created, they are often sourced from existing material that were meant to test humans, with not enough effort put into curating custom benchmarks that can accurately judge a model’s abilities. Finally, in addition to creating and publishing benchmarks for others to use, we believe it is also essential to iron out in detail the entire process of how to create such benchmarks so that it will be easier for future researchers to replicate the process for other domains.

The lack of proper benchmarks, of course, is not without reason. Creating a benchmark is a complicated and time-consuming process, and in fields like science, care needs to be taken to verify the benchmarks are properly balanced across a variety of competing criteria. They need to be balanced for difficulty, usefulness, and accuracy, with the benchmark needing to be challenging enough for current models while also being achievable in the near future, and be a good mix of questions that can truly assess the capabilities while staying within the range of the limits of current systems.

In this paper, we introduce NuclearQA\footnote{The dataset is available at \url{https://github.com/pnnl/EXPERT2}}: a novel, expert-crafted benchmark for evaluating the scientific understanding of large language models in the nuclear domain, encompassing fields like physics, material science, chemistry, etc. Unlike a lot of other benchmarks that use tests made for humans and adapt it for the models, we built our benchmark from scratch exclusively to test scientific understanding of LLMs. We not only present and describe the NuclearQA benchmark, but also lay out in full detail our approach of creating a high quality benchmark that can properly evaluate a model’s scientific understanding. We show how we created a balanced benchmark to be a true test of understanding of nuclear-related science for LLMs. Additionally, we evaluate some of the state-of-the-art models with our questions and observe that even the best LLMs lack scientific knowledge required to excel in our benchmark.

\section{Related Works}

There have been numerous works in the field of question answering for quite some time. While some of them focus on general question-answering abilities of models, others have focused on question answering (QA) of a particular domain.

\subsection{General QA Benchmarks}

There have been numerous benchmarks that deal with the general question-answering abilities of models. Perhaps the most famous is the Stanford Question Answering Dataset (SQuAD)~\cite{rajpurkar2016squad}, consisting of 100,000+ questions and a reading comprehension dataset. They contrast three types of tasks: reading comprehension (RC; read a passage, select a span that answers); Open-domain QA (answer a question from a large set of documents); and Cloze datasets (predict a missing word in a passage). Another pivotal work is the AI2 Reasoning Challenge (ARC)~\cite{clark2018think}. ARC consisted of a dataset of almost $8,000$ science questions in English, and also included a set of questions that neither a retrieval-based algorithm nor a word co-occurrence algorithm were able to answer correctly. Likewise, the MCTest dataset~\cite{richardson2013mctest} consists of a total of $500$ stories and $2000$ multiple-choice reading comprehension questions that were targeted at 7 year olds.

Additionally, there are several other datasets, like CommonsenseQA - 12K multiple-choice questions \cite{talmor2018commonsenseqa}, NewsQA: 10K news articles \cite{trischler2016newsqa}, Search QA: 140K QA pairs \cite{dunn2017searchqa}, TriviaQA: 650K QA pairs with evidence \cite{joshi2017triviaqa}, the ARC2 \cite{bhakthavatsalam2021think}, Big Bench \cite{ghazal2013bigbench}, GLUE \cite{wang2018glue}, and many more that focus on general question-answering abilities.


\subsection{Scientific and Academic Benchmarks}

More recently, there have been several works that focus on using AI models for the scientific domain. As a result, there have been several benchmarks that pertain to this field. Science Questions: 1K multiple choice questions in AI2R \cite{talmor2018commonsenseqa} and SciQ Dataset: \cite{welbl2017crowdsourcing} 13,679 multiple choice science questions are two key and pioneering benchmarks in the scientific domain. Other important works include SciQA \cite{auer2023sciqa}, a benchmark for scientific question answering that was created by using knowledge graphs of academic articles and with the help of human-made templates, and SciRepEval\cite{singh2022scirepeval}, a collection of several scientific document tasks across four types: classification, regression, proximity, and searching. Finally, perhaps one of the most widely used science benchmarks is the science-specific portions of the MMLU \cite{hendrycks2020measuring} benchmark, which include high school and college-level questions for a wide variety of  scientific fields, like Physics, Chemistry, Biology, Computer Science, and many more.

Similarly, some of the other most recent works include QASA \cite{lee2023qasa}, a QA benchmark of \(\sim \)1800 questions to test reasoning on scientific articles, specifically in AI and ML domains, and SciBench\cite{wang2023scibench}, a benchmark of \(\sim \)700 questions sourced from textbooks for college-level science problems. Another recent work in the field is the scientific dataset released by Galactica \cite{taylor2022galactica} alongside their model.

There are also benchmarks that address specific fields, with TheoremQA \cite{chen2023theoremqa} for mathematics, emrQA \cite{pampari2018emrqa} for medicine, and BioRead \cite{pappas2018bioread} and BioMRC \cite{pappas2020biomrc} for biology. 
BigBio~\cite{fries2022bigbio} presents a framework with more than 126 biomedical NLP datasets, and guidelines for task schema, data auditing, etc.

The closest thing to a nuclear benchmark is the NQuAD dataset that was released together with the NukeBERT \cite{jain2020nukebert} model. However, the questions in the NQuAD dataset are selected from pre-sampled paragraphs and contain answers in those specific selection of text. This limits the necessity of a model having to actually understand the nuclear domain, with the ability to comprehend just a small passage of text being sufficient to perform well on this benchmark. In contrast, we include questions in NuclearQA that don't have a specific text containing the answer, but rather needs an understanding of the science to be able to answer correctly. Furthermore, our benchmark has questions across a number of different dimensions. These differences make our benchmark presented here a clear advancement of the work.
\section{The NuclearQA Benchmark}

The NuclearQA benchmark presented in this work is a first-of-its-kind benchmark. It has not been adapted from tests originally meant for humans, but is crafted by subject matter experts (SMEs) specifically to assure that these questions are well suited to judge a language model’s ability to solve nuclear-related questions. While creating this benchmark, we have put every effort into assuring that the benchmark consists of high-quality questions from across disciplines that relate to the nuclear domain, including physics, chemistry, material sciences, and so on. When creating any benchmark, it is important to make sure that the benchmark has a variety of different types of questions such that it can test different types of abilities. As such, NuclearQA has been designed to be balanced across a number of dimensions. We describe the distribution of the questions across these dimensions in detail below.

\subsection{Difficulty}

One of the most natural and important ways to classify the questions is by difficulty. Our benchmark consists of questions of three difficulty levels: \textbf{Easy, Medium,} and \textbf{Hard}, with the questions being divided more or less evenly across the categories. These difficulty levels were defined by SMEs based on the difficulty from a nuclear domain point of view, rather than based on a computational model’s perceived difficulty in solving these questions.

\begin{table}[ht]
\centering
\begin{tabular}{|c|c|}
\hline
\textbf{Difficulty} & \textbf{\% of Questions} \\
\hline
Easy & 31 \\
\hline
Medium & 33 \\
\hline
Hard & 36 \\
\hline
\end{tabular}
\caption{Proportion of the questions for each level of difficulty}
\label{tab:difficulty-stats}
\end{table}

\subsection{Question Format}


The benchmark consists of questions that were considered short-answer questions (\textbf{Short QA}), or more factoid-like in nature, and open-ended long-answer questions (\textbf{Open QA}), which require additional reasoning abilities to answer. Short QA questions are trivia-style questions that can be answered with a few words. The benchmark purposefully favored Short QA, with only a quarter of the questions being Open QA.

\begin{table}[ht]
\centering
\begin{tabular}{|c|c|}
\hline
\textbf{Difficulty} & \textbf{\% of Questions} \\
\hline
Short QA & 75 \\
\hline
Open QA & 25 \\
\hline
\end{tabular}
\caption{Proportion of the questions based on question format}
\label{tab:question-format-stats}
\end{table}

\subsection{Answer Format}

This dimension is based on whether the answer contains a single or a composite correct answer. If the question has a clear single answer, it’s classified as \textbf{single correct}. If it has multiple correct answers that make up a full correct answer, that is classified as \textbf{multiple correct}. For example, for the question \textit{What are the three main subatomic particles?} the full correct answer contains three components, i.e., proton, neutron, and electron. Finally, there are some questions that can’t be put into either of these bins. These are typically the open-ended questions whose answers are open to interpretation. We denote the \textbf{uncategorizable} as \textbf{N/A} in the dataset.

\begin{table}[ht]
\centering
\begin{tabular}{|c|c|}
\hline
\textbf{Answer Format} & \textbf{\% of Questions} \\
\hline
Single Correct & 60 \\
\hline
Multiple Correct & 30 \\
\hline
N/A & 10 \\
\hline
\end{tabular}
\caption{Proportion of questions based on the answer format}
\label{tab:answer-format-stats}
\end{table}

\subsection{Answer Type}

This set of classification has to do with the type of response that would be the correct answer. The dimension is named to be closer to the meaning of \textit{type} in a more programming sense of the word. We have defined four main types: Numerical, Scientific, Numerical + Scientific, and General.

As the name suggests, if the answer is a number, that is classified as \textbf{Numerical}. Questions whose answers have something specifically scientific as response, such as an element symbol or specific quark name, etc., are classified as \textbf{Scientific}. When the answer contains a combination of both, it is classified as \textbf{Numerical + Scientific}. These are answers that require a quantitative and qualitative response. Examples include answers such as \textit{10 protons + 12 neutrons} or \textit{12 moles of Hydrogen}, and so on. Any other question that cannot be categorized as previously described is classified as \textbf{General}. It is important to note that answers to general questions might still include scientific or numerical components, but are not limited to those classifications.

\begin{table}[ht]
\centering
\begin{tabular}{|c|c|}
\hline
\textbf{Type} & \textbf{\% of Questions} \\
\hline
Numerical & 17 \\
\hline
Scientific & 26 \\
\hline
Numerical + Scientific & 20 \\
\hline
General & 37 \\
\hline
\end{tabular}
\caption{Proportion of questions that have answers of a certain type}
\label{tab:answer-type-stats}
\end{table}

\begin{table*}[ht]
\centering
\begin{tabularx}{\textwidth}{|c|X|}
\hline
\textbf{Type} & \textbf{Example question} \\
\hline
\textbf{Numerical} & How many neutrons are inside a U-238 atom? \\
\hline
\textbf{Scientific} & What two particles are emitted after a pair production absorption of a gamma-ray? \\
\hline
\textbf{Numerical + Scientific} & How many Uranium-235 atoms per cubic centimeter are there in natural uranium? \\
\hline
\textbf{General} & Why are poison rods included in some nuclear reactor designs? \\
\hline
\end{tabularx}
\caption{Random examples of questions of different answer types from the NuclearQA dataset}
\end{table*}
\section{Creating the NuclearQA Benchmark}

\subsection{Subject Matter Experts as Question Creators}

One of the first decisions to make when creating the dataset is how to go about creating the questions. While a handful of tools exist that can automatically extract questions from text \cite{cui2021onestop, heilman2011automatic}, we found that none of these questions were of a sufficient quality to be used for evaluating models. We hesitated to use questions that a model can automatically extract as the means to test similar models: we felt this would not be a true test. Using automatic methods would be considerably more economical from both a time and money point of view, but would compromise the quality of the dataset. Thus, we decided that the questions should be curated by humans.

The standard approach of collecting human-written questions for a dataset in cases where existing resources are unavailable is to use some form of crowdsourcing platform  \cite{sap2019atomic, acharya2020towards}. However, given the technical nature of the field, we did not think it would be advisable to have the general public create these questions. We decided that subject matter experts themselves need to create the questions manually to assure quality. One side effect of this was that the total number of questions that could be included in the benchmark would be significantly low compared to what a crowdsourced approach could achieve; on the other hand, the questions themselves would be of the highest possible quality. We decided to pursue quality over quantity.

\subsection{Deciding on Different Types of Questions}

Once we decided on the approach of the benchmark creation, we needed to decide on the different types of questions we wanted to include in the benchmark. The goal of this was to assure we covered a wide breadth of the nuclear domain with some level of depth, while also ensuring it resulted in a useful test for LLMs.

The first thing we wanted in the benchmark was to have questions of varying levels of difficulties so that it could quickly check how models perform compared to each other. We eventually decided on three levels of difficulty. Second, we also wanted to make sure we could test the model with both short-answer questions and open-ended questions in the benchmark. But unlike the difficulty levels that we wanted to distribute more or less evenly, we wanted to assure that we had more short-answer questions than open-ended ones.

Additionally, we wanted to include questions that needed specific scientific answers to be true to the field. We also included some questions that include numerical answers. Eventually, we decided on questions with four different answer types.

Furthermore, because we wanted to see how the models would perform in a format similar to that for a human pupil taking a nuclear sciences exam, we had different types of answers: some only had a single correct answer, some needed multiple correct answers to form a full composite correct response, and some needed reasoning to get to the correct answer. After we decided on these dimensions, we set about creating the questions. We did not set a hard boundary of having a fixed number of questions in each of these categories. Rather, we focused on creating a well-rounded nuclear test with these categories in mind, and made sure to balance them out to reasonable proportions in the end. Through an exhaustive process of checks and edits, we created a benchmark that balanced these categories across several dimensions to the required proportions, as shown in Tables \ref{tab:difficulty-stats}, \ref{tab:question-format-stats}, \ref{tab:answer-format-stats}, and \ref{tab:answer-type-stats}.
\section{Human-in-the-loop Evaluation}

\subsection{Failure of Traditional Metrics}

Due to the nature of our benchmark, traditional methods of evaluation are not suited to judge the success of models on our questions. We have to consider various factors in advance for the selection of good evaluation criteria.

The existing metrics such as partial/exact match accuracy and F1 would not be able to portray an accurate picture of a system’s performance on NuclearQA. For example, if the question was to state the symbol for helium, the answer "H" would be marked a 50\% match by traditional methods, which would of course be completely wrong from a nuclear point of view.

We also experimented with different automatic metrics for different answer types (e.g., numerical, text). However, we noticed that the scale of error is significantly different for atomic numbers and the masses of subatomic particles. For example, an answer of 7 for the atomic number of oxygen is clearly incorrect, while 7.99, which would be essentially 8 from a computational standpoint, is also incorrect because oxygen cannot have a fractional atomic number. Having individual automated metrics to evaluate certain sub-components of the benchmark would introduce a large number of composite metrics, which would be meaningless in terms of the overall performance of the systems.

\subsection{Evaluation Metric and Method}

To alleviate the issues explained in the previous section, we propose a human-in-the-loop evaluation system in place for this benchmark. The first challenge was to come up with a judging criteria with the right scale of evaluation. For example, we did not want to simply have a correct/incorrect categorization, but a scale that is truly reflective of the abilities of LLMs. Thus, we came up with a scale to evaluate the responses, shown in Table \ref{tab:evaluation-criteria}.

\begin{table}[ht]
\centering
\begin{tabular}{|c|c|}
\hline
Score & Meaning \\
\hline
5 & Correct \\
\hline
4 & Partially Correct \\
\hline
3 & Incorrect but related \\
\hline
2 & Unrelated but in-domain \\
\hline
1 & Out-domain and/or nonsensical \\
\hline
\end{tabular}
\caption{Evaluation scale for our human-in-the-loop evaluation}
\label{tab:evaluation-criteria}
\end{table}

\begin{table*}[ht]
\centering
\begin{tabularx}{\textwidth}{|X|X|X|X|X|X|X|}
\hline
& \textbf{Correct} & \textbf{Partially Correct} & \textbf{Incorrect, related} & \textbf{Unrelated, in-domain} & \textbf{Nonsense} & \textbf{Average Score}\\
\hline
\textbf{Llama 2} & \textbf{27} & 10 & 21 & 10 & \textbf{32} & 2.90 \\
\hline
\textbf{Galactica} & 16 & \textbf{13} & 29 & 23 & 19 & 2.84 \\
\hline
\textbf{FlanT5} & 13 & \textbf{13} & \textbf{50} & 18 & 6 & \textbf{3.09} \\
\hline
\textbf{UnifiedQA} & 5 & 4 & 11 & 48 & \textbf{32} & 2.02 \\
\hline
\end{tabularx}
\caption{Olympics-style ratings of the various models' performance on NuclearQA, i.e., models with the highest number of correct answers are shown at the top, regardless of the average score overall, which may be inflated by a lot of relevant but incorrect answers. The best score(s) for each score category are shown in bold. Total number of questions = 100.}
\label{tab:olympic-results}
\end{table*}

We chose different evaluation criteria for short and open question answering (QA). For Short QA, the answer to the corresponding question only has a single correct answer, although the answers can be partially correct and require an SME’s evaluation. For Open QA, an interpretation of the NuclearQA answer is needed, as there is a chance that there is not just one answer to the corresponding question. In Short QA, it requires additional interpretation depending on the number of answers available.

A Short QA evaluation of "5" means that the answer was correct and no interpretation is needed. For multiple answers, a "5" is given if the criteria of the question was met with all correct answers. If an answer is required and was not given, the evaluation was not given a "5." An Open QA evaluation of "5" means that the model provided a correct answer that met the criteria of the question, even if other answers exist. An evaluation of "4" for both types of questions means that the model provided an answer that was partially correct. For multiple answers in Short QA questions, this means that multiple answers are required to be correct but not given (e.g, two correct answers of six total answers). When an answer is provided that is related to the topic of the question but incorrect, that answer is evaluated as a "3." For answers that are unrelated to the question, but still in the general domain of nuclear, the answer is evaluated as a "2." An evaluation of "1" is given to answers that are out of domain or nonsensical. These answers are often related to the model providing an answer in the form of a question, or hallucinating strange text that doesn’t make sense in the context of the question.

\subsection{Baseline Models Evaluation}

When selecting the LLMs to test NuclearQA, we wanted to assure that we selected not just the most popular LLMs, but also the most representative models. We tested the NuclearQA benchmark with four different state-of-the-art LLMs, shown in Table~\ref{tab:eval-model-properties}.

\begin{table}[ht]
\centering
\begin{tabular}{|c|c|c|}
\hline
Model Type & Model & \# of Parameters \\
\hline
\multirow{2}{*}{Encoder-Decoder} & UnifiedQA & 770M \\
\cline{2-3}
 & Flan T5 & 770M \\
\hline
\multirow{2}{*}{Decoder} & Galactica & 1.3B \\
\cline{2-3}
& Llama 2 & 7B \\
\hline
\end{tabular}
\caption{The models and the number of parameters used for evaluation against NuclearQA.}
\label{tab:eval-model-properties}
\end{table}

\begin{enumerate}[i]
    \item \textbf{UnifiedQA} \cite{khashabi2020unifiedqa} is fine-tuned on question-answering datasets, including sets of scientific questions, over the T5 base model.
    \item \textbf{Flan T5} \cite{chung2022scaling} is a instruction-tuned model over the T5 base model.
    \item \textbf{Llama 2} \cite{touvron2023llama} is one of the best performing decoder-style models, exceling in multiple academic benchmarks.
    \item \textbf{Galactica} \cite{taylor2022galactica} is trained with scientific data, including research publications across multiple scientific disciplines.
\end{enumerate}


\begin{figure*}[ht]
\centering
\begin{minipage}{\linewidth}  
\begin{tabular}{p{\linewidth}}  
\hline

\textbf{Question: What are the interaction types of photons with matter?}\\
\textbf{Response:} \\
I've been reading about photons and their interactions with matter. I understand that the photon can be absorbed, reflected, refracted, and transmitted. I'm wondering what the other types of interactions are.\\

Answer: 
A photon can be reflected, absorbed, or transmitted.
A photon can be diffracted.
[...]
A photon can be converted to a phonon.

Comment: This is a good answer, but it is a bit too long. I would recommend to split it into multiple answers.

Comment: @[Username], yes, that is a good idea.

Comment: I've added a few more, but I think this is pretty comprehensive.

Comment: @[Username], I've added a few more.

\hrulefill
\end{tabular}
\end{minipage}
\caption{An example of a response where the model hallucinates a conversation with real people to answer the prompt question. The response has been formatted for clarity and truncated for space. The username has been removed for privacy.}
\label{fig:halluciante-example-1}
\end{figure*}

\subsection{Model Performance}

We used the standard prompting method for all four models with the same configuration setting across all types of questions. We increased the response length to assure the full answer generation for Open QA. The models were not penalized for generating repetitive but correct answers to short questions due to this setting. An SME reviewed the responses for all of these models with no prior knowledge or expectation of which model was expected to perform better or worse to avoid bias.
While we also calculated the average score for all the models, this does not properly represent the overall performance of the models. This is due to the unique nature of our benchmark where many related but incorrect answers could overshadow several completely correct ones. Instead, we used an Olympic medal tally style evaluation, i.e., we treated the model that got the most correct answers as the best model, regardless of the overall average score. However, we also reported the average score for all the models. The full results for the models are shown in Table \ref{tab:olympic-results}.

We can see that the Llama 2 model outperformed the other models by quite a fair distance, getting 27\% of the questions completely correct, with the next best being Galactica with just 16 correct answers. On the other hand, we see that Llama, along with UnifiedQA, also produced the highest number of nonsensical answers. The Flan T5 model managed to produce the highest number of responses that were at least related to the question regardless of correctness, with 76\% of the answers achieving a score of 3 or higher. Flan T5 also produced the fewest number of nonsensical responses, with just 6\% of the responses being nonsensical, less than a third of the next best model, Galactica.
\section{Error Analysis}

While most of the responses don't need a lot of further analysis and are simply incorrect answers, we saw some unique responses from some of these models that require a further look.

We have seen in the past that large language models are prone to hallucinations \cite{rawte2023survey, ji2023survey}, and there have been several efforts to detect and mitigate these hallucinations \cite{manakul2023selfcheckgpt, li2023halueval, zhang2023mitigating}
. In our evaluation, the Llama model seemed the most likely to hallucinate information in the responses, with it constantly making up many of its own questions in the response, and answering those questions. There were also several instances of it generating the responses in the form of a chat between two or more people. These instances could either have been a direct copy from the training data, suggesting the model is memorizing the training data, or else they could be hallucinations. Either way, we found out that the usernames Llama used for these responses were actual usernames of real people on Twitter, and so we have chosen not to disclose those responses verbatim in this paper. A sample of an anonymized version of such response is shown in Figure~\ref{fig:halluciante-example-1}. Furthermore, the Llama model also had the habit of hallucinating its own multiple choice answers for the prompted question and selecting one of them as the answer, with several instances of all its manufactured options being the incorrect answer, and sometimes all the manufactured options being the same one repeated multiple times.

Additionally, the Galactica model sometimes had issues of creating its own question unrelated to the prompt question and then going on to solve its own questions instead. It also had the issue of hallucinating its own multiple-choice answers like the Llama model. With the Flan T5 model, there were a couple of cases with the model producing an empty response. The UnifiedQA model had the least amount of such issues, but there were a couple of instances where the model simply extracted keywords from the questions that happened to be close enough to the correct answer. 

\section{Limitations and Future Work}

While our work in this paper has achieved the goal of creating a novel and comprehensive benchmark, there is still room for further development and refinement. Our main limitation is that this approach requires an extensive time commitment from an SME and therefore is costly to build large datasets. If we are to scale this work to thousands of questions, there would need to be an automated step to speed up the question creation without compromising the quality. Similarly, another limitation is the lack of relevant automated evaluation metrics in the literature for us to use. This is a big gap in the field that needs to be filled if we want to have true measures of success of large language models moving forward.

One obvious way this work could be expanded is by adding more questions across several other domains. Another potential direction for the future is to create queries of other types and not be limited to just the question-answer format.
\section{Conclusion}

In this work, we presented a novel benchmark that is able to accurately evaluate a large language model’s understanding of the nuclear domain. In addition, we laid out in detail the methodology of creating a scientific benchmark, which can serve well for future researchers to use when creating similar benchmarks in other scientific domains. Our results suggest that while the current state-of-the- art LLMs seem to perform the best as expected in the general domain, there is a lot of room for improvement when it comes to demonstrating truly good performance in the cross-disciplinary science domains. Thus, we see that due to its unique nature and quality and variety of questions, NuclearQA is an appropriate measure of a model’s understanding of the nuclear domain and therefore a true test for any such models in the future.
\section*{Acknowledgments}

This work was supported by the NNSA Office of Defense Nuclear Nonproliferation Research and Development, U.S. Department of Energy, and Pacific Northwest National Laboratory, which is operated by Battelle Memorial Institute for the U.S. Department of Energy under Contract DE-AC05–76RLO1830. This article has been cleared by PNNL for public release as PNNL-SA-190898.


\bibliography{anthology,custom}


\end{document}